\RequirePackage[T1]{fontenc}
\documentclass[journal,10pt,letterpaper]{IEEEtran}

\usepackage[utf8]{inputenc}
\usepackage{graphicx}
\usepackage{amsmath,amssymb}
\usepackage{booktabs}
\usepackage[table]{xcolor}
\usepackage{array}
\usepackage{multirow}
\usepackage{makecell}
\usepackage{microtype}
\usepackage{cite}
\usepackage[colorlinks=true,linkcolor=blue!60!black,citecolor=black,urlcolor=blue!60!black,hypertexnames=false]{hyperref}
\usepackage[caption=false,font=footnotesize]{subfig}
\usepackage[ruled,linesnumbered,vlined]{algorithm2e}
\SetAlFnt{\small}
\SetAlCapFnt{\small}
\SetAlCapNameFnt{\small}

\usepackage{tabularx}
\usepackage{siunitx}
\newcolumntype{R}{>{\raggedleft\arraybackslash}X}
\usepackage{url}
\ifdefined\pdfminorversion
\else
  \AtBeginDvi{}
\fi
\hypersetup{
  pdftitle={Response-Oriented Prompt Selection with SAM 3 for Open-Vocabulary Remote-Sensing Segmentation},
  pdfauthor={Yanghui Song, Chengfu Yang, Yongpeng Wen, Zhanxuan Hu, Wei Lu, Qi Ming, Nanqing Liu, Heng-Chao Li},
  pdfsubject={IEEE Transactions on Geoscience and Remote Sensing (TGRS)},
  pdflang={en-US}
}
\usepackage{algpseudocode}
\usepackage{float}
\usepackage{etoolbox}

\newcommand{\citep}{\cite}
\newcommand{\method}{ROSE-SAM3}
\newcommand{\cmark}{\ensuremath{\checkmark}}
\newcommand{\best}[1]{\textbf{#1}}
\newcommand{\secondbest}[1]{\underline{#1}}
\definecolor{tableheader}{gray}{0.90}
\definecolor{tablesection}{gray}{0.95}
\definecolor{tableblue}{RGB}{225,242,239}

\begin{document}

\title{ROSE-SAM3: Response-Oriented Prompt Selection for Open-Vocabulary Remote Sensing Segmentation}

\author{Yanghui Song, Chengfu Yang, Yongpeng Wen, Zhanxuan Hu,\\
Wei Lu, Qi Ming, Nanqing Liu, and Heng-Chao Li,~\IEEEmembership{Senior~Member,~IEEE}%
\thanks{This work has been submitted to the IEEE for possible publication. Copyright may be transferred without notice, after which this version may no longer be accessible.}%
\thanks{This work was supported in part by the National Natural Science Foundation of China under Grant 62601704, and in part by the Basic Research Project of Natural Science of Yunnan Province under Grant 202601AT070017. (Corresponding author: Nanqing Liu)}%
\thanks{Yanghui Song, Chengfu Yang, Yongpeng Wen, Zhanxuan Hu, and Nanqing Liu are with the School of Information Science and Technology, Yunnan Normal University, Kunming, China (e-mail: yanghuisong55@gmail.com; yangchengfu@ynnu.edu.cn; yongpengwen11@163.com; zhanxuanhu@gmail.com; lansing163@163.com).}%
\thanks{Wei Lu is with the School of Computer Science and Technology, Anhui University, Hefei, China (e-mail: luwei@ahu.edu.cn).}%
\thanks{Qi Ming is with the College of Computer Science, Beijing University of Technology, Beijing, China (e-mail: chaser.ming@gmail.com).}%
\thanks{Heng-Chao Li is with the School of Information Science and Technology, Southwest Jiaotong University, Chengdu 611756, China (e-mail: lihengchao\_78@163.com).}%
}

\markboth{Manuscript Prepared for Submission}%
{Song \MakeLowercase{\textit{et al.}}: Response-Oriented Prompt Selection With SAM~3}

\maketitle

\begin{abstract}
In open-vocabulary semantic segmentation (OVSS) of remote-sensing imagery, class-name prompts can miss objects with varied appearances, while richer descriptions can activate competing classes. We propose Response-Oriented Prompt Selection for SAM~3 (ROSE-SAM3), a training-free framework that selects descriptions using frozen SAM~3 responses on unlabeled target-domain images. We screen candidate descriptions for coverage beyond the native prompts while protecting reliable native class responses, retaining candidates that pass these checks on at least four of five image folds. An offline directed acyclic graph (DAG) search then evaluates candidate combinations and builds a compact PromptBank for each class. Each bank retains the native prompts and adds selected descriptions, which we call residual prompts. At inference, the bank is fixed, and Semantic--Instance Cooperative Head (SICH) uses instance evidence to refine semantic responses before assigning pixel labels. We also introduce RS-OVSS-C to evaluate robustness under six image corruption families across eight remote-sensing datasets, preserving their class vocabularies and reference labels. ROSE-SAM3 achieves 57.3\% mean mIoU across eight clean benchmarks and 54.7\% on RS-OVSS-C, achieving state-of-the-art performance among training-free methods. The code is publicly available at
\url{https://github.com/YanghuiSong/ROSE-SAM3}.
\end{abstract}


\section{Introduction}

Remote-sensing segmentation supports tasks such as land-cover mapping, building extraction, and aerial object analysis, with category definitions varying across datasets and applications \citep{wang2021loveda,xia2023openearthmap,zamir2019isaid,lyu2020uavid}. Conventional models learn classifiers for a fixed category set, limiting their reuse when these definitions change. Open-vocabulary semantic segmentation (OVSS) offers a flexible alternative by specifying target categories through text \citep{radford2021learning,li2025segearth,zermatten2025tacoss}. With a suitable pretrained model, this formulation allows users to query different categories without training a new classifier for each task.

Training-free OVSS further reduces adaptation costs by keeping pretrained model parameters unchanged on the target domain. CLIP-based approaches refine dense image--text correspondence \citep{wang2024sclip,lan2024proxyclip,li2025segearth}, while SAM~3 \citep{carion2025sam} supports segmentation of text-defined concepts. Recent studies extend this capability to remote-sensing imagery \citep{li2025segearthov3,pei2026taming,tien2026activesam,liu2026glopsam3,wang2026rsgp}. In this setting, the choice of class descriptions directly affects segmentation: even with a fixed model, different descriptions of the same category can produce different spatial responses.
This distinction separates mask quality from category assignment: a model may produce a plausible region while assigning it inconsistently across related text concepts. Text-conditioned responses therefore offer a way to inspect both the location and class evidence of a prediction. Our method uses these responses to select descriptions before inference, complementing approaches that refine masks or combine geometric cues after prompts have been chosen.

\begin{figure*}[!t]
    \centering
    \subfloat[\label{fig:prompt_motivation_a}]{%
        \includegraphics[
            height=0.265\textwidth,
            keepaspectratio
        ]{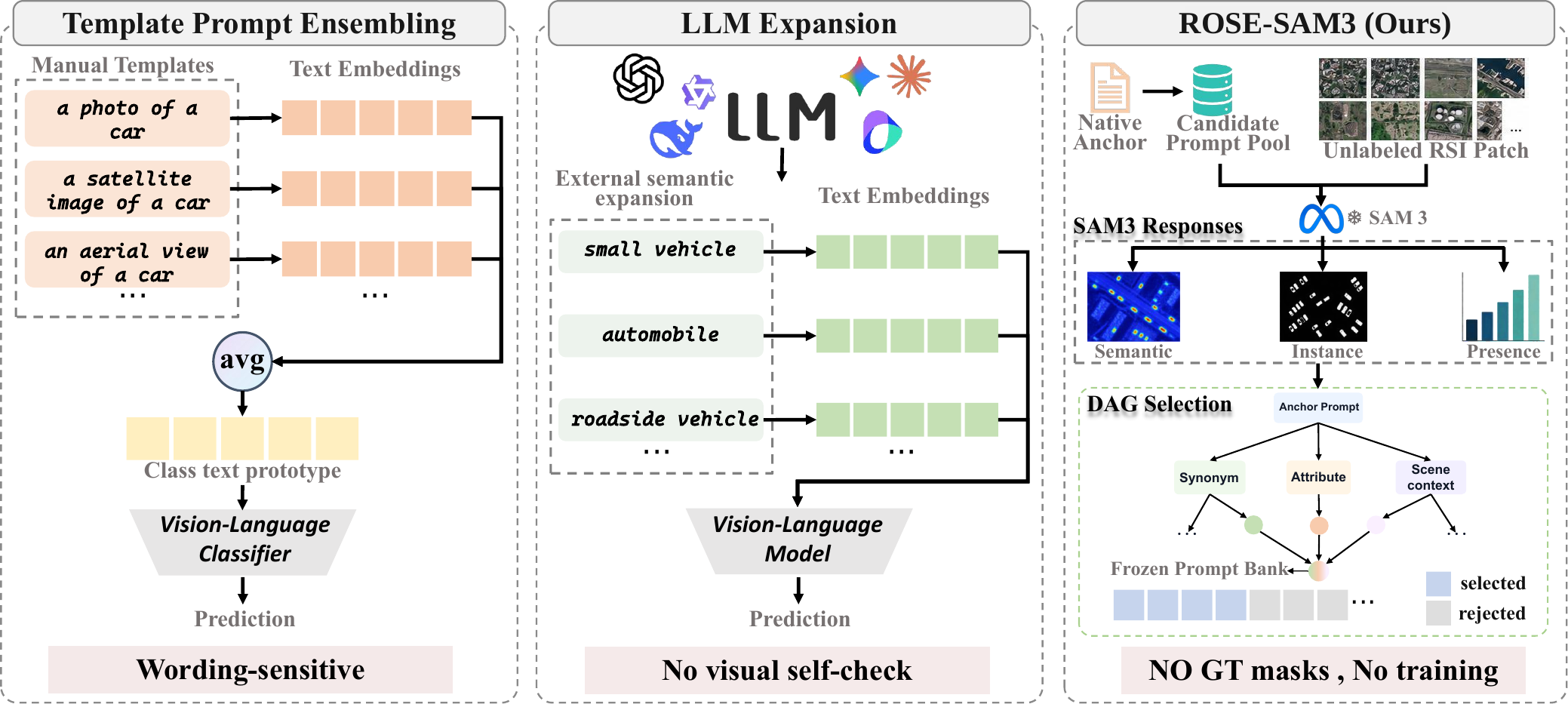}%
    }
    \hspace{0.01\textwidth}
    \subfloat[\label{fig:prompt_motivation_b}]{%
        \includegraphics[
            height=0.265\textwidth,
            keepaspectratio
        ]{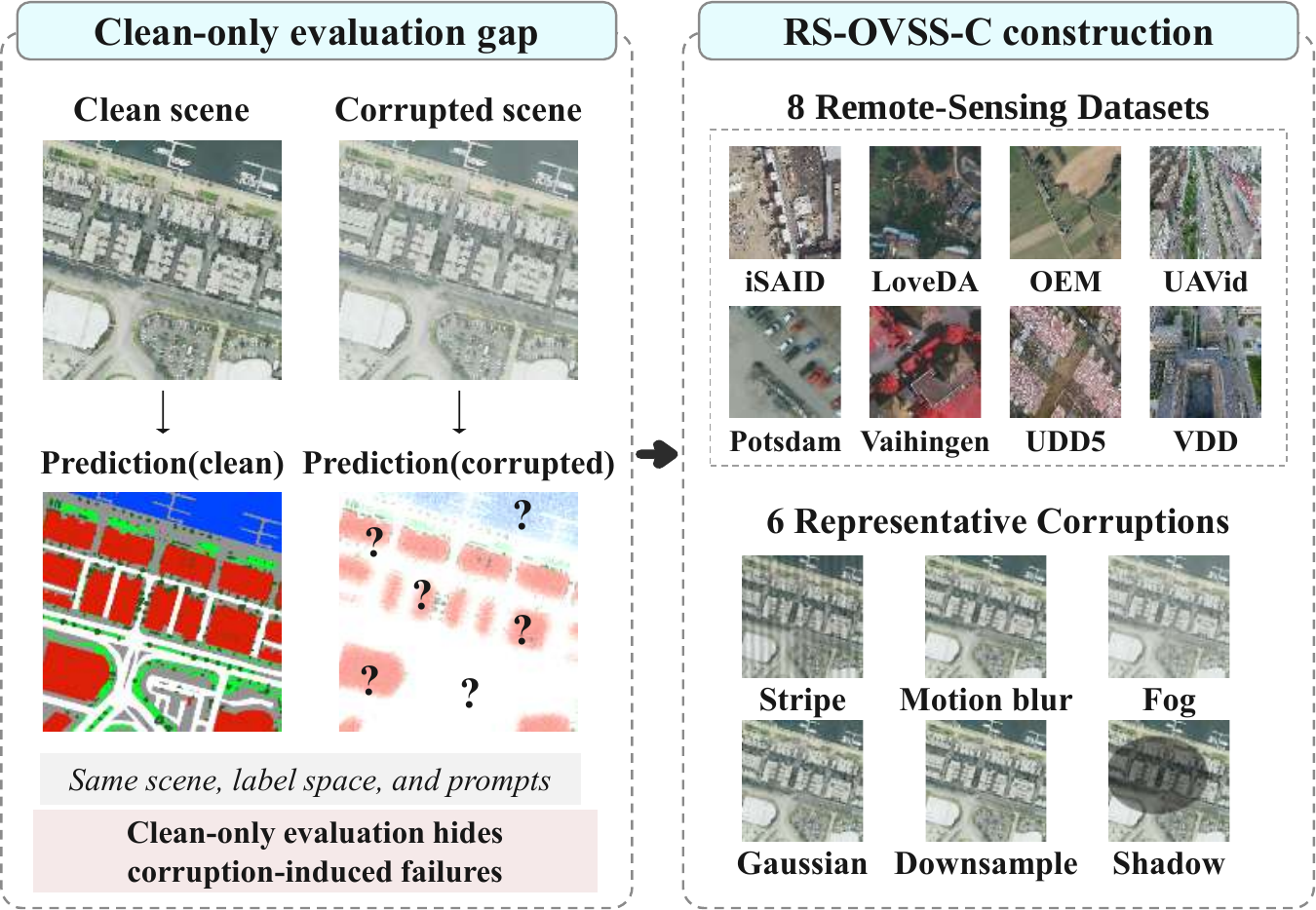}%
    }

    \caption{Motivation for ROSE-SAM3. Prompt descriptions can be
    assessed through their spatial responses, while controlled
    corruptions expose errors that clean-image evaluation misses.}

    \label{fig:prompt_motivation}
\end{figure*}

For each class, a fixed set of basic class descriptions forms the native prompt set, which serves as the anchor for selection. These prompts can miss regions with varied appearances, motivating richer descriptions through template ensembling or language-model expansion \citep{pratt2023cupl,lin2023clipes,benigmim2025floss}. Fig.~\ref{fig:prompt_motivation}(a) contrasts these strategies with our approach. Template ensembling averages the text embeddings of predefined descriptions into a class prototype, while LLM-based expansion generates a broader set of descriptions. A semantically relevant description, however, may simply repeat the anchor's response or activate competing classes. Responses on target images reveal these overlaps, even though their accuracy cannot be measured without annotations. The challenge is to use this unlabeled evidence to select descriptions that cover additional target regions while limiting cross-class interference.

We address this challenge with \textbf{R}esponse-\textbf{O}riented Prompt \textbf{Se}lection for \textbf{SAM~3} (\textbf{ROSE-SAM3}), a training-free framework that selects prompts using frozen SAM~3 responses on unlabeled target-domain images. Image evidence has also been used for class purification and concept selection \citep{chen2025freecp,chen2026toward}. We focus on choosing complementary descriptions for each dataset class and fixing them before inference. Agreement between semantic and instance responses provides evidence that a region supports the prompted concept. Checks across image folds test whether a candidate contributes consistently across images. We also measure coverage beyond the native anchor and reject candidates that override consistently predicted regions of other classes. These checks require no segmentation labels, but individually useful candidates may still be redundant when combined.

We therefore evaluate candidate combinations using their combined responses. A small prompt budget limits the search, which we organize as a directed acyclic graph (DAG) rooted at the native anchor. Each node represents a prompt set, and each edge adds one candidate. The search produces a class-specific PromptBank containing the native anchor and selected additional descriptions, termed residual prompts. This bank remains fixed for all query images. Semantic--Instance Cooperative Head (SICH) then uses instance evidence to refine each prompt's semantic response before class prediction. SAM~3 parameters remain frozen throughout selection and inference.

Clean-image evaluation alone does not show how segmentation methods handle noise, blur, and other image degradations, as illustrated in Fig.~\ref{fig:prompt_motivation}(b). We introduce RS-OVSS-C, a controlled robustness benchmark that applies six corruption families to eight remote-sensing datasets. It preserves class vocabularies, reference labels, and evaluation protocols so that methods can be compared under the same degradations.

The main contributions are:
\begin{itemize}
\item We propose \textbf{ROSE-SAM3}. Offline DAG selection uses frozen responses on unlabeled target-domain images to construct fixed class-specific PromptBanks without training or ground-truth segmentation annotations.

\item We introduce RS-OVSS-C, a controlled benchmark evaluating segmentation robustness across six corruption families and eight remote-sensing datasets while preserving class vocabularies and reference labels.

\item Experiments show gains over native anchors and competitive training-free baselines on clean and corrupted images. Ablations assess DAG selection, SICH, and prompt budgets.

\end{itemize}

\section{Related Work}

\subsection{Open-Vocabulary Segmentation for Remote Sensing}

Open-vocabulary semantic segmentation connects dense image predictions to categories specified through text. Early approaches establish this connection through language-guided masks, region grouping, and dense feature correspondence \citep{ghiasi2022scaling,li2022language,xu2022groupvit,xu2023san,cho2024catseg}. Subsequent work improves visual--textual alignment through learned adaptation or distillation. LDVC \citep{zhang2025ldvc} and GBA \citep{xu2025gba} use visual consensus and correlation constraints, respectively, while CLIP-to-Seg \citep{chen2026cliptoseg} distills vision--language knowledge and FOV-Seg \citep{chng2026fovseg} learns fine-grained word--region alignment. Training-free approaches pursue dense correspondence by refining pretrained features without updating model parameters \citep{liu2024scan,wang2024sclip,lan2024proxyclip,zhao2025dpseg}.

Remote-sensing images vary in spatial resolution, viewing conditions, scene composition, and category definitions \citep{zhan2023rsvg}, with cross-scene variability also investigated in hyperspectral coastal wetland mapping \citep{Hu2026HSIFrFRWKV}. Research addresses these conditions through remote-sensing vision--language models, including RemoteCLIP, GeoChat, and SkySense \citep{liu2024remoteclip,kuckreja2024geochat,guo2024skysense,zhan2025skyeyegpt}, and segmentation approaches such as OVRS and SegEarth-OV \citep{cao2024ovrs,li2025segearth,li2026efficient}, as well as selective representation learning from foundation vision models in EVSR \citep{Ma2026EVSR}. Related studies further investigate prompt adaptation, modality alignment, and semantic ambiguity in OVSS \citep{an2025softguided,zhang2025tpovseg,zavras2025modalitygap,huang2026kgovrseg}, alongside structure-constrained semantic contrastive learning for referring remote-sensing image segmentation \citep{Lei2026S2CLNet}. These approaches improve visual features or their alignment with text, but the spatial prediction also depends on how each category is described. This is especially relevant when a class contains objects with varied appearances or its name has a similar meaning to another class name.

Remote-sensing OVSS must also handle large changes in object scale and background composition within the same category. A single label may cover both compact objects and broad land-cover regions, while visually similar classes can occur side by side. These properties make prompt wording consequential even when the image and text encoders are fixed. ROSE-SAM3 addresses this prompt-level source of variation by comparing candidate descriptions through their predicted regions on target-domain images.

\subsection{Segment Anything Models for Remote Sensing}

The Segment Anything Model (SAM) \citep{kirillov2023segment} predicts class-agnostic masks from spatial prompts, supporting remote-sensing annotation, dataset construction, and auxiliary segmentation \citep{wang2023samrs,ma2024samassisted,zhang2025alps,liu2025pointsam}, while SAM-based multimodal fine-tuning has also been explored for remote sensing semantic segmentation \citep{Ma2025MultimodalFineTuning}. Its transfer to remote-sensing imagery is affected by differences in object scale, resolution, and scene content \citep{osco2023samrs,ren2024segment}. Moreover, class-agnostic mask generation alone does not assign regions to dataset categories, so semantic segmentation requires additional information for class recognition. SAM~3 \citep{carion2025sam} incorporates text-defined concepts into promptable segmentation and provides semantic, instance, and presence responses. Recent remote-sensing methods use this interface to refine semantic predictions or provide geometric cues \citep{li2025segearthov3,song2026prompt,wang2026rsgp,liu2026glopsam3}. Even after mask refinement, their predictions can vary with the wording of the text prompt.

\begin{figure*}[!t]
    \centering
    \includegraphics[
        width=0.98\linewidth,
        height=0.78\textheight,
        keepaspectratio
    ]{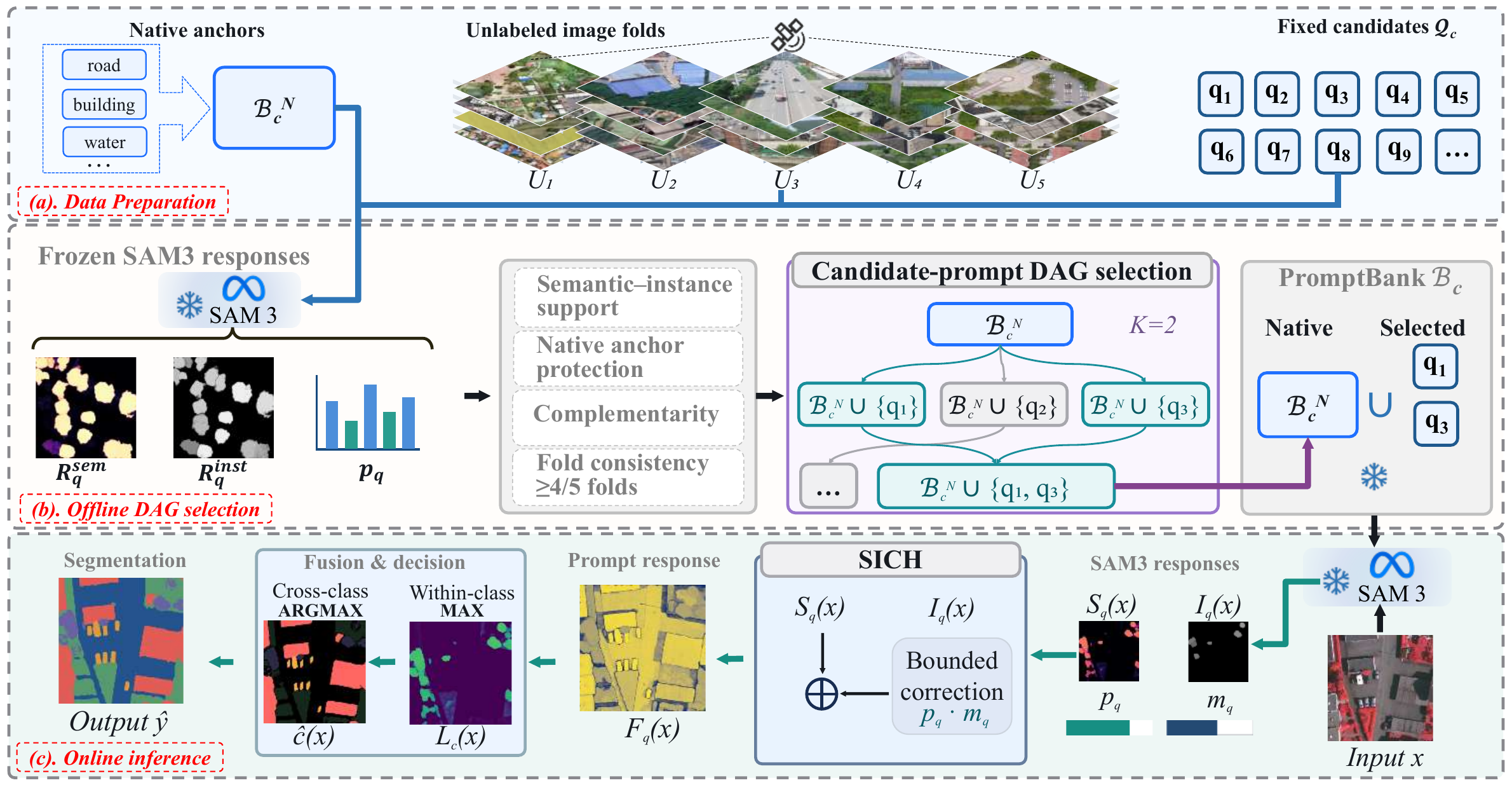}
    \par\vspace{-4pt}

    \caption{ROSE-SAM3 pipeline. Native anchors and candidate
    descriptions are evaluated using frozen SAM~3 responses on
    unlabeled image folds. Offline DAG selection constructs a
    class-specific PromptBank, which SICH uses for online segmentation.}
    \label{fig:method}
\end{figure*}

\subsection{Prompt Construction and Selection}

Prompt construction expands class names through templates, synonyms, or generated descriptions. CLIP \citep{radford2021learning} combines class names with predefined templates, while CuPL \citep{pratt2023cupl} uses a language model to generate class-specific descriptions. For segmentation, CLIP-ES, ITACLIP, and FLOSS investigate textual guidance and class-dependent prompting \citep{lin2023clipes,aydin2025itaclip,benigmim2025floss}. Richer descriptions also introduce a selection problem, since additional text can provide redundant or irrelevant information. BBN \citep{pan2025purify} addresses noise in text embeddings before visual feature aggregation. These operations refine text representations without directly testing whether a description covers additional regions of a class.

Model responses can also guide selection. FreeCP \citep{chen2025freecp} uses spatial activations to reduce redundant categories for each image, while CG-ICS \citep{chen2026toward} selects generated concepts using SAM~3 scores and reference examples within an in-context episode. These methods select categories or concepts for a given image or episode. Our task is to select a description set that can be reused across images for a fixed dataset vocabulary. This distinction affects what evidence the selection procedure must aggregate. Per-image selection can adapt to the current scene, whereas a reusable PromptBank must provide consistent coverage across different scenes without seeing their labels. ROSE-SAM3 evaluates candidate responses across image folds and checks prompt combinations, so the selected descriptions are fixed for subsequent images and do not require a separate selection step for each query.

\subsection{Robustness under Image Corruptions}

Robustness benchmarks complement clean-image evaluation by measuring performance under degraded image conditions. Common-corruption benchmarks introduce controlled noise, blur, weather effects, and digital artifacts \citep{hendrycks2019benchmarking}, with subsequent work extending this evaluation to semantic segmentation \citep{kamann2021corruptions}. For promptable segmentation, Qiao et al.\ \citep{qiao2023robustness} evaluate SAM under common corruptions and style changes, as well as local occlusion and adversarial attacks. Beyond controlled perturbations, ACDC \citep{sakaridis2021acdc} examines real driving scenes under fog, nighttime, rain, and snow. In remote sensing, robustness studies cover land-cover classification under fog \citep{shi2023semanticfog}, continual test-time adaptation for detection on DIOR-C \citep{yong2026ctaod}, and Earth-observation models under image perturbations in REOBench \citep{li2025reobench}. Recent OVSS benchmarks also broaden dataset and task coverage \citep{li2026realistic}. These studies span different prediction tasks and degradation protocols. For remote-sensing OVSS, a controlled test with the same class vocabulary and reference labels across clean and corrupted images makes changes in segmentation performance directly comparable.

Such evaluation also tests whether text-conditioned responses remain useful when visual evidence is weakened. A prompt selected on clean images may still recover a class under noise or fog, while corruption can remove fine structures needed to distinguish neighboring regions. RS-OVSS-C measures both outcomes with fixed labels and matched image degradations.

\section{Methodology}
\label{sec:method}

\subsection{Overview}

ROSE-SAM3 selects a compact set of descriptions for each dataset
class using SAM~3 responses on unlabeled target-domain images.
The goal is to cover regions missed by the native anchors while
preserving regions consistently assigned to other classes.
As shown in Fig.~\ref{fig:method}, we prepare native prompts
and candidate descriptions for each class and obtain their responses
on unlabeled calibration images.
We first screen individual descriptions for complementary coverage
while preserving reliable predictions from the native prompts.
We then organize the retained descriptions into a directed
acyclic graph (DAG), where each node represents a prompt combination
and each edge adds one description.
The DAG search selects a complementary prompt set under a fixed budget.
The selected descriptions are combined with the native prompts to
form a PromptBank for each class.

During online inference, SAM~3 processes each test image using the
class-specific PromptBanks constructed offline.
The Semantic--Instance Cooperative Head (SICH) combines SAM~3's
semantic and instance predictions to refine the response to each
prompt. We aggregate the refined responses within each class and
assign each pixel to the class with the highest score.
Prompt selection requires no
ground-truth segmentation annotations, and SAM~3 parameters remain
frozen during both selection and inference.

\subsection{Data Preparation}
\label{sec:candidate_generation}

Let $\mathcal{C}=\{c_k\}_{k=1}^{N_{\mathcal C}}$ denote the target dataset's class vocabulary. For each class $c\in\mathcal C$, the native prompt set $\mathcal{B}^{N}_{c}$ contains basic descriptions of that class, as shown in Fig.~\ref{fig:method}(a). These prompts are retained throughout selection. We use $S_{\mathcal{B}^{N}_{c}}$ to denote their aggregated semantic response. The candidate set $\mathcal{Q}_{c}=\{q_{c,1},q_{c,2},\ldots,q_{c,n_c}\}$ contains additional descriptions that the search may select.
We prepare candidate descriptions based on the visual characteristics of each class. This preparation uses neither SAM~3 responses nor image annotations. The resulting candidates and native prompts are fixed before DAG selection.

To check whether candidate descriptions behave consistently across images, we divide the unlabeled calibration images into five folds, as shown in Fig.~\ref{fig:method}(a). Candidate screening evaluates each description against the native anchor, and the bounded DAG search then evaluates combinations of the candidates that pass screening. Both stages use the same cached responses on these folds. We cache responses for each calibration image and its horizontally flipped copy, aligning the latter to the original image to assess prediction consistency.

\subsection{Offline DAG Selection}
\label{sec:dag_selection}

\begin{figure}[!t]
    \centering
    \includegraphics[width=\columnwidth]
    {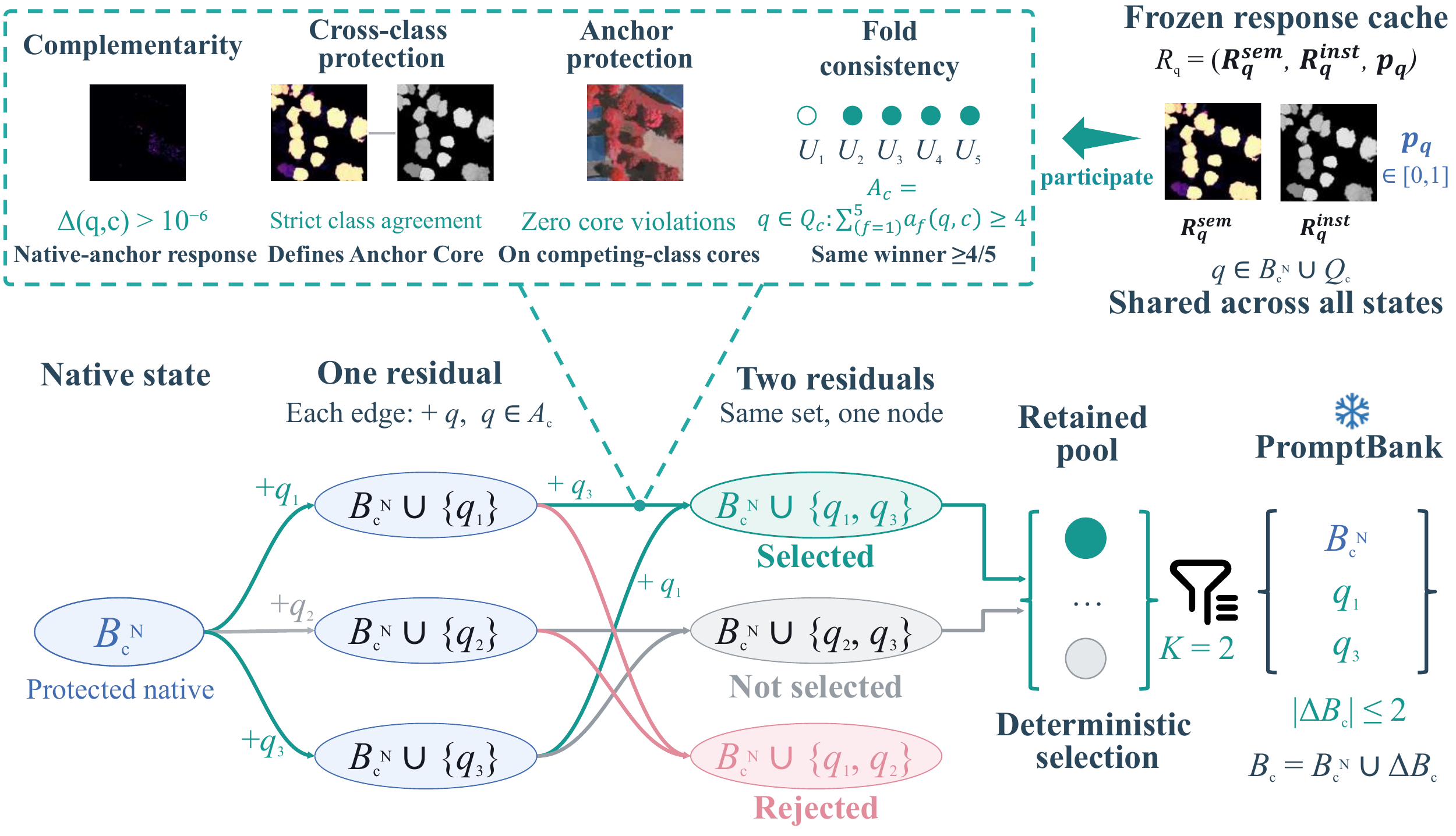}
    \caption{Candidate screening and bounded DAG selection. The
    response checks identify useful descriptions while protecting
    native class responses. The DAG then evaluates their combinations
    and retains a compact PromptBank.}
    \label{fig:dag_selection}
\end{figure}

Offline selection first identifies descriptions that complement the
native prompts, then tests whether they remain useful when combined.
Both stages use cached SAM~3 responses on unlabeled calibration images,
as shown in Fig.~\ref{fig:dag_selection}. For every native prompt and
candidate $q\in\mathcal{B}^{N}_{c}\cup\mathcal{Q}_{c}$, we cache its
semantic response $S_q$, instance response $I_q$, and presence score
$p_q$. The native responses $S_{\mathcal{B}^{N}_{c}}$ and
$I_{\mathcal{B}^{N}_{c}}$ are obtained by pixel-wise maximization over
native prompts. We write $S_q(i,x)$ for the semantic response at pixel
$x$ in image $i$, and use the fused selection response
$\widetilde R_q(i,x)=p_q(i)\max\{S_q(i,x),I_q(i,x)\}$.
These cached responses support both selection stages without updating
SAM~3 or consulting segmentation labels.

\subsubsection{Candidate Screening}

Screening retains descriptions that activate regions missed
by native prompts without overriding reliable competing-class
predictions. For example, a rooftop description may recover missed
buildings but should not relabel a consistently identified road.
For class $c$ and fold $f$, we define the native-supported region:
\begin{equation}
\mathcal{A}^{N}_{c,f}=\{(i,x):S_{\mathcal{B}^{N}_{c}}(i,x)>\tau_N\},
\label{eq:native_region}
\end{equation}
where $\tau_N$ is fixed. We measure additional response outside
this region:
\begin{equation}
\Delta_f(q,c)=\operatorname{Agg}_{(i,x)\notin\mathcal{A}^{N}_{c,f}}
\bigl[\widetilde R_q(i,x)-\widetilde R_{\mathcal{B}^{N}_{c}}(i,x)\bigr]_+.
\label{eq:complementarity}
\end{equation}
Here $[u]_+=\max(u,0)$, $\widetilde R_q$ is the fused candidate
response, and $\widetilde R_{\mathcal{B}^{N}_{c}}$ is its aggregated
native counterpart. The fixed operator $\operatorname{Agg}$ summarizes
positive gains across the fold; we also count images and pixels with
gains.

For cross-class protection, we identify native anchor cores where
semantic, instance, and fused class responses share the same unique
highest-scoring class. For a candidate $q$ of class $c$, $V_f(q,c)$
counts pixels in competing-class cores where its response exceeds
the native winning response. Thus, agreement defines protected
regions, while anchor protection prevents their replacement.
A candidate passes fold $f$ if $\Delta_f(q,c)>10^{-6}$ and
$V_f(q,c)=0$. The screened set $\mathcal{A}_c$ contains candidates
passing at least four of five folds. Native prompts remain fixed.

\begin{figure}[!t]
    \centering
    \includegraphics[width=\columnwidth]
    {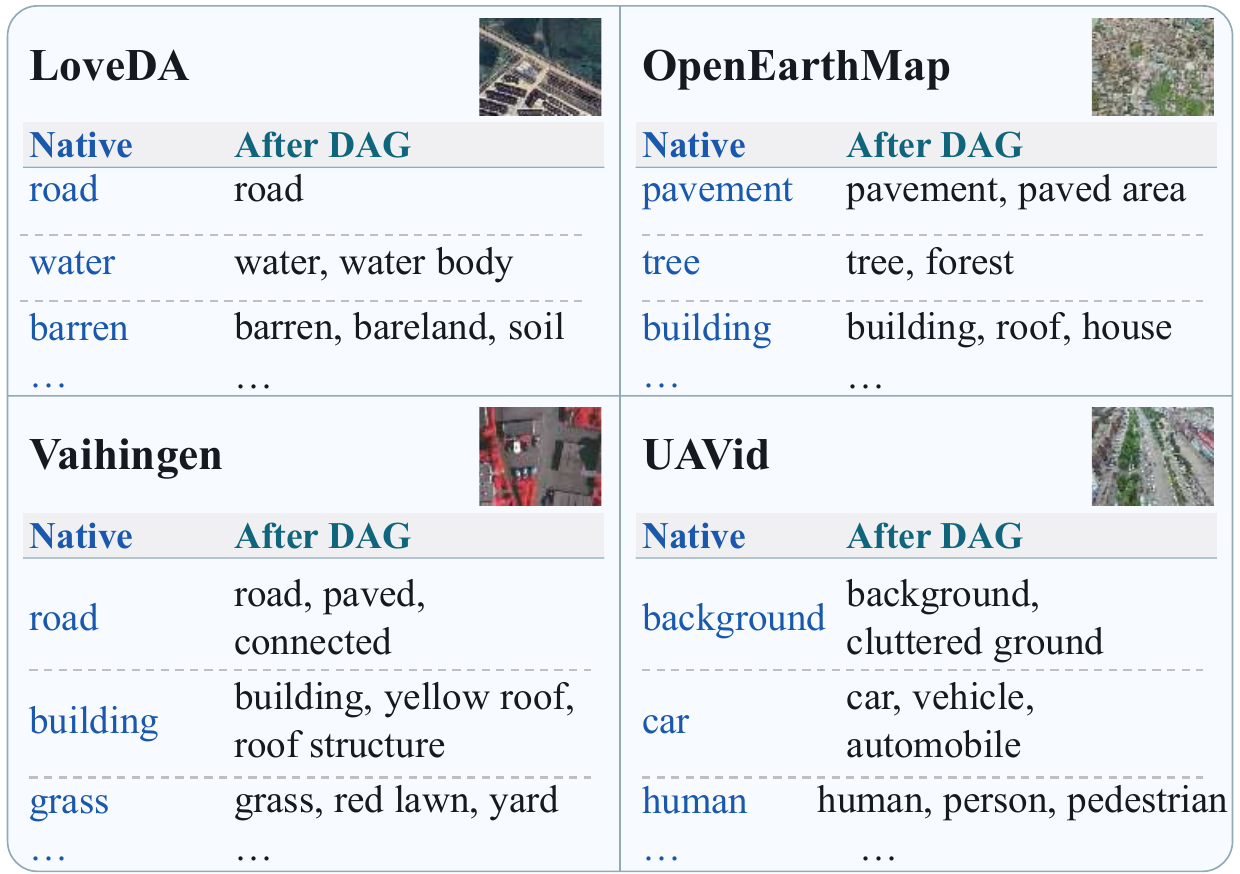}
    \caption{Examples of native anchors and residual prompts retained
    after DAG selection. The added descriptions express alternative
    appearances or names for each dataset class.}
    \label{fig:dag_prompt_vocabulary}
\end{figure}

\subsubsection{Bounded DAG Search}

Individual screening cannot account for overlap or interference
between accepted descriptions. Candidates useful in isolation may be
redundant together or change neighboring class predictions after
aggregation. We therefore evaluate prompt
combinations in a bounded directed acyclic graph (DAG), rooted at
the native set $\mathcal{B}^{N}_{c}$. Each node represents a prompt
set and each edge adds one screened description from $\mathcal{A}_{c}$.
Layer $k$ contains states with $k$ additional prompts, subject to
budget $K$. Fig.~\ref{fig:dag_selection} illustrates $K=2$:
\begin{equation}
\mathcal{V}^{\mathrm{sel}}_{c}
=
\bigl\{
\mathcal{B}^{N}_{c}\cup\mathcal{D}:
\mathcal{D}\subseteq\mathcal{A}_{c},
\ |\mathcal{D}|\leq K
\bigr\}.
\label{eq:selection_state_dag}
\end{equation}
Different insertion orders yielding the same set lead to one node.
For example, adding $q_1$ then $q_3$, or $q_3$ then $q_1$,
produces the same state in Fig.~\ref{fig:dag_selection}. Each state
is evaluated once using cached responses, without further SAM~3
queries. Only the combination changes; the model and cached response
maps remain fixed during the search.

For each state, we aggregate its prompt responses and compare the
result with the native reference. The combined response must add
coverage without violating protected cores of competing classes.
This check is necessary because individually accepted prompts can
interfere after aggregation. Protection is reassessed on the combined
response, rather than assumed from individual candidate decisions.
In Fig.~\ref{fig:dag_selection}, the
$\{q_1,q_2\}$ state is rejected, whereas $\{q_2,q_3\}$ is feasible
but not selected. We exclude states violating the fixed constraints,
then rank admissible states on each fold by preserved native support
and complementary coverage. Fixed identifiers break ties. The same
state must win on at least four folds; otherwise, selection falls
back to the native set. The winner forms the fixed PromptBank:
\begin{equation}
\mathcal{B}_{c}=\mathcal{B}^{N}_{c}\cup\Delta\mathcal{B}_{c},
\qquad |\Delta\mathcal{B}_{c}|\leq K.
\label{eq:final_promptbank}
\end{equation}
Fig.~\ref{fig:dag_prompt_vocabulary} shows examples of retained
descriptions, while Algorithm~\ref{alg:dag_prompt_selection}
summarizes the offline procedure.

\begin{algorithm}[!t]
\caption{Offline DAG Selection}
\label{alg:dag_prompt_selection}

\KwIn{Cached frozen SAM~3 responses, five unlabeled
selection folds $\{\mathcal{U}_{f}\}_{f=1}^{5}$,
native prompt sets $\{\mathcal{B}^{N}_{c}\}$, candidate sets
$\{\mathcal{Q}_{c}\}$, and active budget $K$}

\KwOut{Frozen PromptBanks $\{\mathcal{B}_{c}\}$}

\For{each class $c\in\mathcal{C}$}{
    Start from the native prompt set
    $\mathcal{B}^{N}_{c}$\;

    \tcp{Candidate screening}
    \For{each candidate $q\in\mathcal{Q}_{c}$}{
        \For{each fold $f$}{
            Compute complementarity
            $\Delta_f(q,c)$\;

            Evaluate core violations
            $V_f(q,c)$\;

            $a_f(q,c)\gets1$ iff
            $\Delta_f(q,c)>10^{-6}$
            and $V_f(q,c)=0$\;
        }
    }

    $\mathcal{A}_{c}\gets
    \{q:\sum_{f=1}^{5}a_f(q,c)\geq4\}$\;

    \tcp{Bounded DAG construction}
    Construct states
    $\mathcal{B}^{N}_{c}\cup\mathcal{D}$,
    where $\mathcal{D}\subseteq\mathcal{A}_{c}$
    and $|\mathcal{D}|\leq K$\;

    \tcp{Fold-consistent state selection}
    \For{each fold $f$}{
        Evaluate states using cached responses\;
        Exclude states violating the fixed constraints\;
        Select the deterministic winning state
        $\widehat{\mathcal{B}}_{c,f}$\;
    }

    Select the same winning state if it wins
    on at least four folds. Otherwise retain
    $\mathcal{B}^{N}_{c}$\;

    Freeze the selected PromptBank $\mathcal{B}_{c}$\;
}

\Return{$\{\mathcal{B}_{c}\}$}\;
\end{algorithm}

\subsection{Online Inference}
\label{sec:response_fusion}

\begin{figure}[!t]
    \centering
    \includegraphics[width=\columnwidth]
    {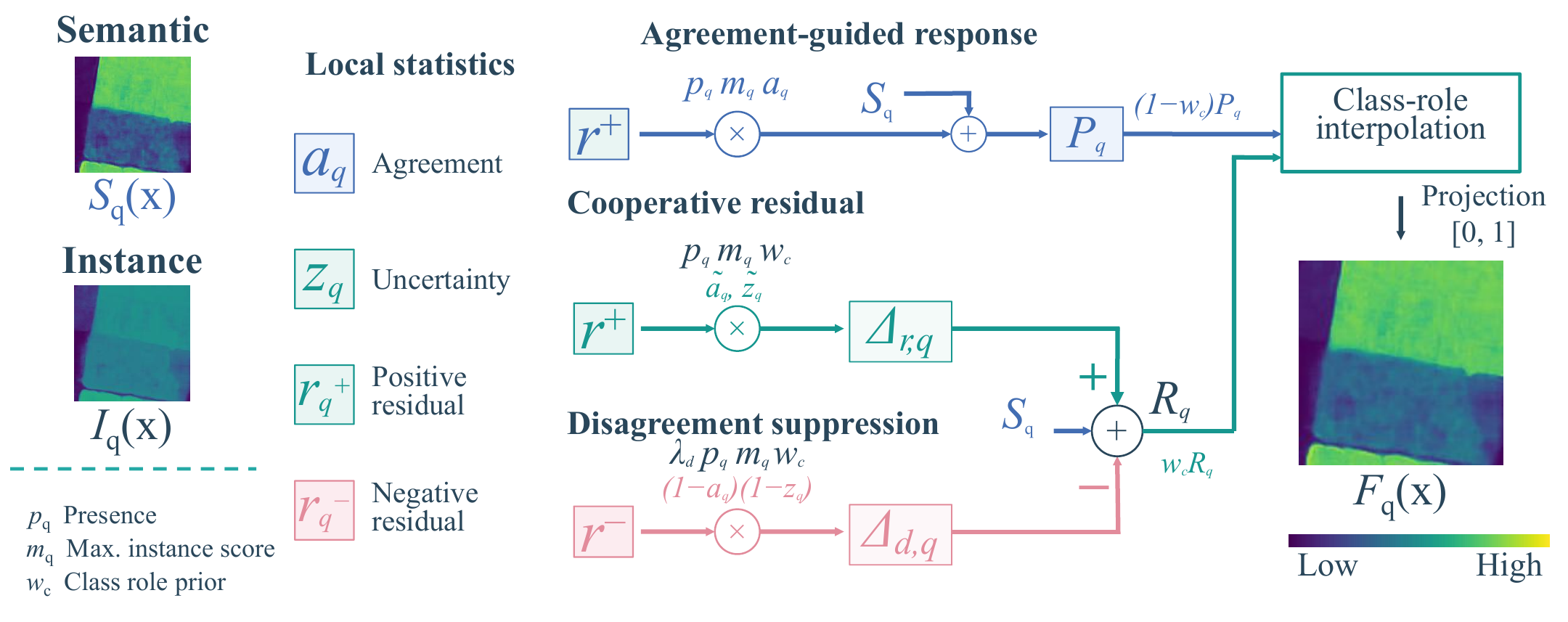}
    \par\vspace{-6pt}
    \caption{Semantic--Instance Cooperative Head (SICH).
    Instance evidence reinforces or suppresses semantic responses.
    A fixed class weight combines the two refinement branches.}
    \label{fig:sich_overview}
\end{figure}

Once DAG selection has fixed the PromptBank, each query image is
processed with those descriptions. For a prompt $q$ in class $c$,
frozen SAM~3 returns semantic and instance responses $S_q(x)$ and
$I_q(x)$ at pixel $x$. It also returns a presence score $p_q$ for
the prompted concept and the highest valid instance score $m_q$.
SICH uses the instance evidence to adjust the semantic response.
Fig.~\ref{fig:sich_overview} traces this process from the two
response maps to the refined prompt response $F_q$.

The adjustment depends on how the maps compare at each pixel.
Local agreement $a_q=1-|S_q-I_q|$ is high when they are close,
while uncertainty $z_q=1-2|S_q-0.5|$ is high when the semantic
response is near $0.5$. We separate the two directions of
disagreement: $r_q^+=[I_q-S_q]_+$ marks instance support beyond
the semantic response, and $r_q^-=[S_q-I_q]_+$ marks semantic
activation beyond the instance response. Here $[u]_+=\max(u,0)$,
and we omit the pixel argument $x$ for readability.
SICH uses two branches to control added instance support and
suppression of unsupported semantic activation separately:
\begin{equation}
\begin{aligned}
P_q(x)&=S_q(x)+p_qm_q a_q(x)r_q^+(x),\\
R_q(x)&=S_q(x)+\Delta_{r,q}(x)-\Delta_{d,q}(x).
\end{aligned}
\label{eq:sich_branches}
\end{equation}
The first branch, $P_q$, can only raise the semantic response
where the instance response is stronger. Presence $p_q$ and
instance score $m_q$ limit that increase when prompt-level
support is weak. The second branch, $R_q$, can both reinforce
and suppress the semantic response through two corrections:
\begin{equation}
\begin{aligned}
\Delta_{r,q}(x)
&=p_qm_qw_c\,
\tilde a_q(x)\tilde z_q(x)r_q^+(x),\\
\Delta_{d,q}(x)
&=\lambda_d p_qm_qw_c
(1-a_q(x))(1-z_q(x))r_q^-(x),
\end{aligned}
\label{eq:sich_corrections}
\end{equation}
Here $w_c\in[0,1]$ is a fixed weight for class $c$, and
$\lambda_d$ controls suppression. The factors
$\tilde a_q=\rho+(1-\rho)a_q$ and
$\tilde z_q=\rho+(1-\rho)z_q$ have a minimum value of
$\rho\in[0,1]$, so reinforcement is not eliminated when agreement
or semantic uncertainty is low. Stronger instance evidence can recover a
region with a weak semantic response, while semantic activation
with little instance support can be reduced when disagreement is clear.

The class weight $w_c$ combines the branches into one response:
\begin{equation}
F_q(x)=\Pi_{[0,1]}\left(
(1-w_c)P_q(x)+w_cR_q(x)
\right),
\label{eq:sich_response}
\end{equation}
where $\Pi_{[0,1]}$ clips values to $[0,1]$. At $w_c=0$,
only $P_q$ contributes. At $w_c=1$, $R_q$ determines the output.
For each class, we keep its strongest prompt response at each
pixel, $L_c(x)=\max_{q\in\mathcal{B}_c}F_q(x)$. The class with
the highest score is predicted if that score reaches
$\tau_{\mathrm{prob}}$. Otherwise, the pixel is background.
All correction coefficients, class weights,
and the background threshold are fixed before
evaluation.

\section{Experiments}
\subsection{Datasets and Corruption Benchmark}

\begin{figure}[!t]
    \centering
    \includegraphics[width=\columnwidth]{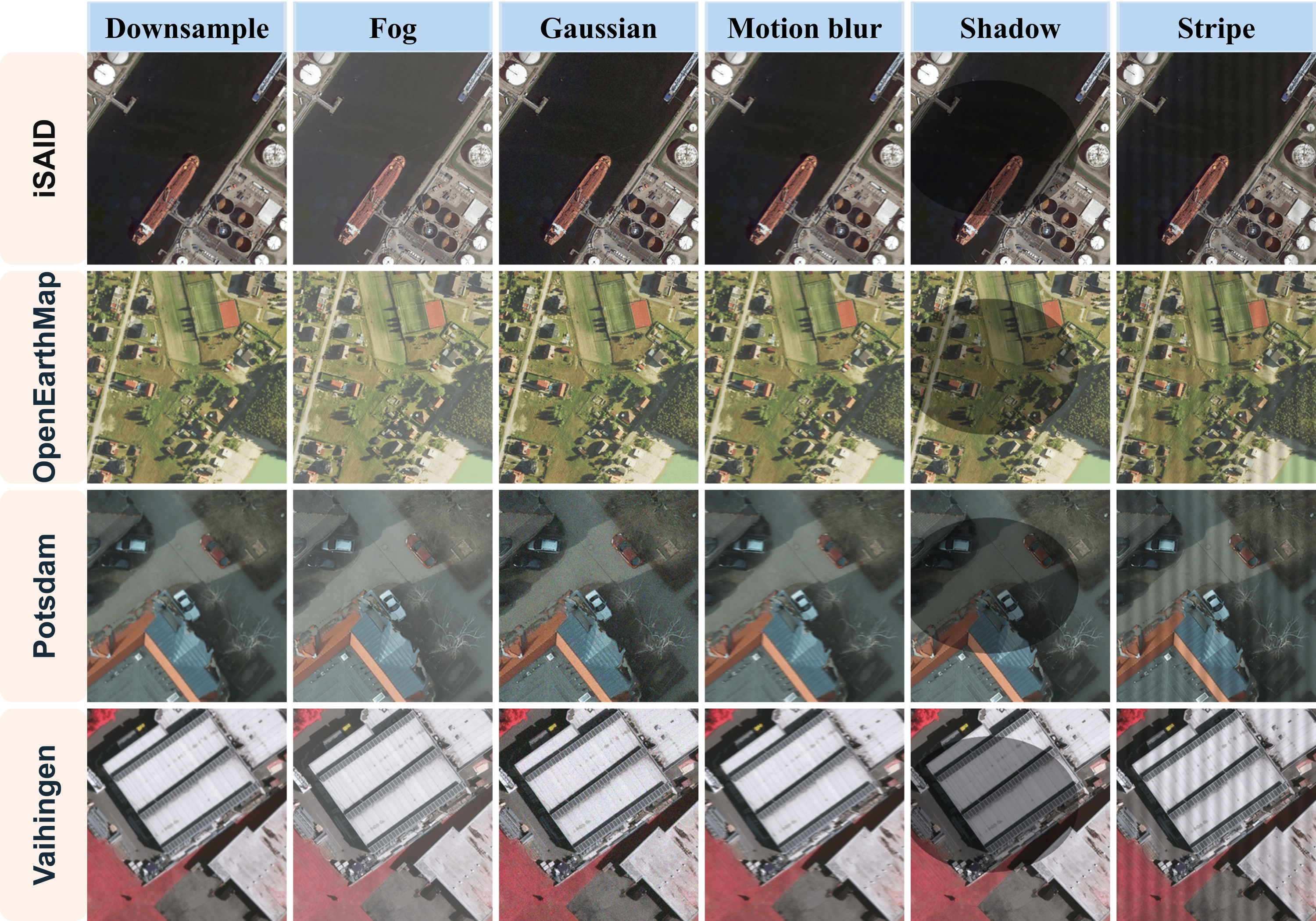}
    \par\vspace{-7pt}
    \caption{RS-OVSS-C examples under the reported corruption settings.
    The corruptions change image appearance while preserving image
    dimensions and segmentation labels.}
    \label{fig:corruption_examples}
\end{figure}

The clean-image evaluation and RS-OVSS-C use the same eight source datasets: OpenEarthMap (OEM)~\citep{xia2023openearthmap}, LoveDA~\citep{wang2021loveda}, iSAID~\citep{zamir2019isaid}, Potsdam~\citep{isprs_potsdam}, Vaihingen~\citep{isprs_vaihingen}, UAVid~\citep{lyu2020uavid}, UDD5~\citep{chen2018udd}, and VDD~\citep{cai2025vdd}. These datasets cover land-cover mapping, aerial object segmentation, and UAV scenes at different spatial resolutions. We construct RS-OVSS-C by applying controlled image corruptions to each dataset, allowing robustness to be assessed alongside clean-image performance. Fig.~\ref{fig:corruption_examples} shows six corruption families that affect pixel values, spatial detail, contrast, and illumination: Gaussian noise, motion blur, downsampling, fog, shadow, and stripe artifacts. Each corruption is applied separately to the source RGB image. The corrupted image retains the spatial dimensions and pixel
alignment of its clean counterpart, while the class vocabulary
and ground-truth segmentation mask remain unchanged throughout
evaluation.

The reported evaluation uses one fixed setting for each corruption.
\textbf{Gaussian noise} adds independent zero-mean Gaussian perturbations
to the RGB pixel values. \textbf{Motion blur} averages pixels along a
horizontal window, using edge padding to preserve image dimensions.
\textbf{Downsampling} first reduces the image resolution and then
restores its original dimensions using bilinear interpolation for
both operations. This removes fine spatial detail without changing
the label grid. \textbf{Fog} blends the source image with a uniform
light-gray image whose RGB values are all 235, reducing contrast
through a spatially uniform veil. \textbf{Shadow} multiplies pixel
intensities inside a fixed elliptical region by an attenuation
factor, leaving pixels outside the region unchanged.
\textbf{Stripe artifacts} multiply pixel intensities by a periodic
sinusoidal profile along the horizontal axis, producing vertical
bands of varying brightness. Pixel values are clipped to $[0,255]$
and converted to 8-bit RGB where needed. For Gaussian noise, the
random seed is determined by a base seed and the image path, so the
same image and configuration produce the same corrupted input.
Exact parameter values and generation scripts are available in the
released code.

\begin{table*}[!t]
\caption{Clean-image open-vocabulary segmentation comparison (mIoU, \%). $^{*}$ marks results reproduced from public implementations. Other values are from the cited studies.}
\label{tab:overall_clean_reconstructed}
\centering
\scriptsize
\renewcommand{\arraystretch}{0.94}
\setlength{\tabcolsep}{0.75pt}
\begin{tabular*}{\textwidth}{@{\extracolsep{\fill}}ll*{9}{>{\footnotesize}c}@{}}
\toprule
\textbf{Method} & \textbf{Backbone} & \textbf{LoveDA} & \textbf{Potsdam} & \textbf{Vaihingen} & \textbf{iSAID} & \textbf{OEM} & \textbf{UDD5} & \textbf{VDD} & \textbf{UAVid} & \textbf{Avg.} \\
\midrule
\rowcolor{tablesection}
\multicolumn{11}{@{}l}{\textit{Trained on remote-sensing segmentation data}} \\
SCAN \citep{liu2024scan} & CLIP-L/14@336 & 23.2 & 27.5 & 15.2 & 44.3 & -- & 34.1 & 29.2 & 20.3 & 27.7 \\
SAN \citep{xu2023san} & CLIP-L/14@336 & 25.3 & 37.3 & 39.2 & 49.6 & -- & 37.2 & 35.8 & 23.5 & 35.4 \\
SED \citep{xie2024sed} & CLIP ConvNeXt-L & 24.6 & 29.4 & 39.0 & 51.2 & -- & 35.7 & 32.5 & 21.3 & 33.4 \\
CAT-Seg \citep{cho2024catseg} & CLIP-L/14@336 & 28.6 & 35.8 & 42.3 & 53.3 & -- & 40.2 & 39.1 & 25.7 & 37.9 \\
OVRS \citep{cao2024ovrs} & CLIP-L/14@336 & 31.5 & 36.4 & 43.5 & 52.7 & -- & 40.8 & 37.2 & 24.1 & 38.0 \\
GSNet \citep{ye2025towards} & CLIP-L/14 + RS-DINO-B/32 & 32.5 & 37.9 & 44.1 & 53.7 & -- & 40.9 & 37.3 & 24.2 & 38.7 \\
RSKT-Seg \citep{li2026efficient} & RemoteCLIP-L/14 + RS-DINO-B/32 & 33.2 & 38.4 & 42.7 & 54.3 & -- & 42.1 & 39.7 & 25.7 & 39.4 \\
SkySense-O \citep{zhu2025skysenseo} & SkySense + CLIP + SAM & 38.3 & 54.1 & 51.6 & 43.9 & 40.8 & -- & -- & -- & 45.7 \\
\midrule
\rowcolor{tablesection}
\multicolumn{11}{@{}l}{\textit{Test-time adaptation}} \\
TMPA \citep{yang2026tmpa} & CLIP-B/16 & 39.7 & 51.1 & 43.4 & 26.2 & 42.2 & 52.4 & 49.0 & 45.9 & 43.7 \\
\midrule
\rowcolor{tablesection}
\multicolumn{11}{@{}l}{\textit{Training-free methods}} \\
CLIP \citep{radford2021learning} & CLIP-B/16 & 12.4 & 15.6 & 10.8 & 7.5 & 12.0 & 9.5 & 14.2 & 10.9 & 11.6 \\
MaskCLIP \citep{ding2023open} & CLIP-B/16 & 27.8 & 33.9 & 29.9 & 14.5 & 25.1 & 32.4 & 32.9 & 28.6 & 28.1 \\
VIP$^{*}$ \citep{zhu2026vip} & DINOv3-L/16 + DINO.txt & 31.2 & 49.8 & 47.0 & 26.1 & 20.7 & 35.2 & 54.3 & 19.9 & 35.5 \\
SCLIP \citep{wang2024sclip} & CLIP-B/16 & 30.4 & 39.6 & 35.9 & 16.1 & 29.3 & 38.7 & 37.9 & 31.4 & 32.4 \\
ClearCLIP \citep{lan2024clearclip} & CLIP-B/16 & 32.4 & 42.0 & 36.2 & 18.2 & 31.0 & 41.8 & 39.3 & 36.2 & 34.6 \\
GEM \citep{bousselham2024grounding} & CLIP-B/16 & 31.6 & 39.1 & 36.4 & 17.7 & 33.9 & 41.2 & 39.5 & 33.4 & 34.1 \\
SegEarth-OV \citep{li2025segearth} & CLIP-B/16 + DINO-B/8 & 36.9 & 48.5 & 29.1 & 21.7 & 40.3 & 50.6 & 45.3 & 42.5 & 39.4 \\
ProxyCLIP \citep{lan2024proxyclip} & CLIP-B/16 + DINOv2-B/14 & 34.3 & 49.0 & 47.5 & 21.8 & 38.9 & 40.8 & 47.8 & 35.8 & 39.5 \\
RSCLIP$^{*}$ \citep{wang2026rsclip} & CLIP-B/16 + DINO-B/8 & 35.5 & 47.4 & 40.3 & 21.2 & 39.8 & 50.4 & 42.5 & 42.5 & 40.0 \\
CorrCLIP \citep{zhang2025corrclip} & MetaCLIP-B/16 + DINO-B/8 & 36.9 & 51.9 & 47.0 & 25.5 & 32.9 & 46.1 & 47.3 & 38.3 & 40.7 \\
ConInfer \citep{chen2026coninfer} & CLIP-B/16 + DINOv3-L/16 & 39.3 & 50.0 & 31.4 & 20.1 & 42.0 & 46.9 & 50.3 & 46.4 & 40.8 \\
ReAttnCLIP \citep{niu2026reattnclip} & CLIP-B/16 & 37.0 & 48.7 & 29.9 & 23.2 & 41.1 & 53.7 & 49.7 & 44.0 & 40.9 \\
\midrule
\rowcolor{tablesection}
\multicolumn{11}{@{}l}{\textit{SAM~3-based training-free methods}} \\
ActiveSAM$^{*}$ \citep{tien2026activesam} & SAM~3 & 41.1 & 40.6 & 45.1 & 28.3 & 40.7 & 72.1 & 62.2 & 53.3 & 47.9 \\
SegEarth-OV3$^{*}$ \citep{li2025segearthov3}
& SAM~3
& 47.4
& 57.9
& 60.8
& 27.6
& 44.2
& 71.7
& 64.5
& 56.8
& 53.8 \\

RSGPNet$^{*}$ \citep{wang2026rsgp}
& SAM~3
& 47.4
& 57.8
& 60.8
& 27.6
& 41.9
& 71.7
& \secondbest{67.3}
& 56.8
& 53.9 \\

GLoP-SAM3 \citep{liu2026glopsam3}
& SAM~3
& 46.2
& 57.9
& 62.2
& 28.0
& 46.6
& 72.6
& 64.7
& 55.4
& 54.2 \\

GFMR \citep{wang2026gfmr}
& SAM~3
& \secondbest{48.3}
& 58.4
& 64.3
& \secondbest{29.9}
& \best{49.5}
& 72.7
& 66.2
& 55.3
& 55.6 \\

ProC-SAM3 \citep{song2026prompt} & SAM~3 & 46.1 & \secondbest{58.8} & \secondbest{64.5} & 29.5 & 48.9 & \secondbest{73.4} & \best{67.5} & \best{60.0} & \secondbest{56.1} \\
\rowcolor{tableblue}
\rlap{\color{tableblue}\rule[-\dp\strutbox]{\textwidth}{\dimexpr\ht\strutbox+\dp\strutbox\relax}}\textbf{\method{} (Ours)} & \textbf{SAM~3} & \best{48.7} & \best{58.9} & \best{65.3} & \best{35.1} & \secondbest{49.3} & \best{73.7} & \best{67.5} & \secondbest{59.9} & \best{57.3} \\
\bottomrule
\end{tabular*}
\end{table*}

\begin{figure*}[!t]
\centering
\includegraphics[width=0.99\linewidth,keepaspectratio]{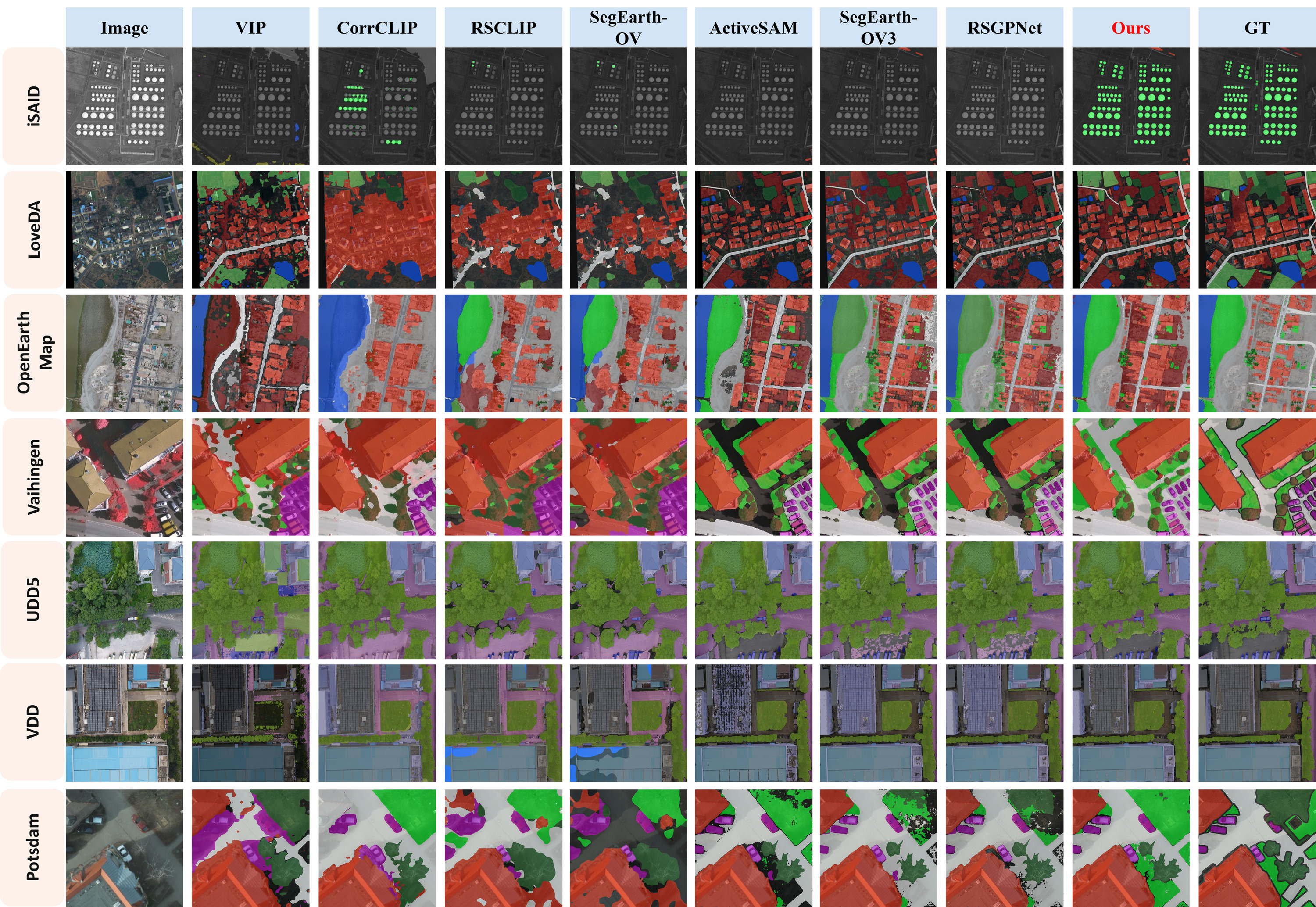}
\par\vspace{-5pt}
\caption{Qualitative comparison on clean remote-sensing images
spanning aerial objects, land-cover regions, and UAV scenes.}
\label{fig:qualitative_clean}
\end{figure*}

\subsection{Experimental Setup}
\label{sec:metrics}
For each dataset, we build the PromptBank from its class vocabulary
and unlabeled calibration images. The calibration images are split
into five non-overlapping folds for DAG selection. For OEM, we use
validation images without their segmentation labels. Test images
are excluded from selection, and the PromptBank is fixed before
evaluating either clean or corrupted images.

We report mean intersection over union (mIoU), computed as the
macro-average of class-wise IoU using each dataset's evaluator.
For RS-OVSS-C, we also report mean accuracy (mAcc), the average
of class-wise pixel accuracy. Results are rounded to one decimal
place unless stated otherwise. In controlled
comparisons, the SAM~3 checkpoint, data split, evaluator, background
rule, and post-processing are held fixed. Dataset-specific decision
thresholds are unchanged between clean and corrupted images.
All experiments run on one NVIDIA GeForce RTX~4090 GPU with 24~GB
of memory. The image encoder uses bf16 precision, while response
fusion uses fp32. SICH coefficients are shared across the eight
datasets.

\subsection{Main Results on Clean Images}
Table~\ref{tab:overall_clean_reconstructed} compares methods on eight clean benchmarks. It includes models trained on labeled remote-sensing segmentation data \citep{cho2024catseg,cao2024ovrs,zhu2025skysenseo}, a test-time adaptation method \citep{yang2026tmpa}, and training-free approaches \citep{wang2024sclip,li2025segearth}. The trained methods adapt dense vision--language features or use remote-sensing-specific models to predict the dataset classes. TMPA updates its predictions at test time using target images, whereas the training-free methods reuse pretrained models without target-domain training. Among the latter methods evaluated on all eight datasets, ROSE-SAM3 achieves the highest mean mIoU at 57.3\%. It ranks first or ties for first on six datasets and places second on OEM and UAVid. We focus below on training-free methods built on SAM~3 \citep{tien2026activesam,li2025segearthov3}, which share our backbone and require no target-domain training.

ROSE-SAM3 exceeds ProC-SAM3~\citep{song2026prompt} and GFMR~\citep{wang2026gfmr} by 1.2 and 1.7 percentage points in mean mIoU, respectively. The largest difference is on iSAID, where it reaches 35.1\%, compared with 29.5\% for ProC-SAM3 and 29.9\% for GFMR. Differences on the other datasets are smaller, with GFMR leading on OEM and ProC-SAM3 on UAVid. Against the reproduced SegEarth-OV3~\citep{li2025segearthov3} baseline, ROSE-SAM3 improves mIoU on all eight datasets, increasing the mean from 53.8\% to 57.3\%. The largest gains occur on iSAID, OEM, and Vaihingen.

Relative to SegEarth-OV3, the gains range from 1.0 points on Potsdam to 7.5 on iSAID, with improvements of 5.1 on OEM and 4.5 on Vaihingen. This variation indicates that prompt selection has different effects across datasets. Fig.~\ref{fig:qualitative_clean} illustrates two relevant error types: missed target regions and confusion between adjacent classes.
ROSE-SAM3 recovers more small circular objects in iSAID and more complete building regions in LoveDA and OEM, while preserving gaps between neighboring targets. It also maintains road continuity in UDD5 and separates the large building from adjacent pavement in Vaihingen. These examples illustrate how selected descriptions can improve target coverage without merging neighboring regions. Thin structures, small objects, and boundaries between visually similar classes remain challenging.

\subsection{Ablation Studies}
\label{sec:ablation}

\paragraph{DAG Selection and SICH}
Table~\ref{tab:component_ablation} evaluates the contributions of
DAG-based prompt selection and Semantic--Instance Cooperative Head
(SICH) on OEM, Potsdam, Vaihingen, and UAVid. The first two
configurations use identical max fusion, allowing the effect of
replacing native anchors with the selected PromptBank to be examined
separately. DAG selection improves mean mIoU from 55.6\% to 58.2\%,
with gains of 1.3, 1.5, 4.5, and 2.8 percentage points on the four
datasets, respectively. The largest gain occurs on Vaihingen, where
the selected descriptions substantially improve performance relative
to the native prompts. With the PromptBank fixed, SICH further raises
the mean to 58.4\%, providing improvements of 0.1--0.3 points across
the datasets. These results show that prompt selection accounts for
most of the observed improvement, while response refinement yields
a smaller additional benefit. The two components operate at distinct
stages: DAG selection determines which descriptions are retained,
whereas SICH adjusts the responses obtained from those descriptions.

\begin{table}[!t]
\caption{Component ablation (mIoU, \%).}
\label{tab:component_ablation}
\centering
\footnotesize
\renewcommand{\arraystretch}{1.08}
\setlength{\tabcolsep}{2pt}
\begin{tabular*}{\columnwidth}
{@{\extracolsep{\fill}}ccrrrrr@{}}
\toprule
\makecell{DAG\\selection} & SICH
& OEM & Potsdam & Vaihingen & UAVid & Mean \\
\midrule
$-$ & $-$
& 47.9 & 57.1 & 60.7 & 56.8 & 55.6 \\
\cmark & $-$
& 49.2 & 58.6 & 65.2 & 59.6 & 58.2 \\
\cmark & \cmark
& \textbf{49.3} & \textbf{58.9}
& \textbf{65.3} & \textbf{59.9}
& \textbf{58.4} \\
\bottomrule
\end{tabular*}
\end{table}

\begin{figure}[!t]
\centering
\includegraphics[width=\linewidth]
{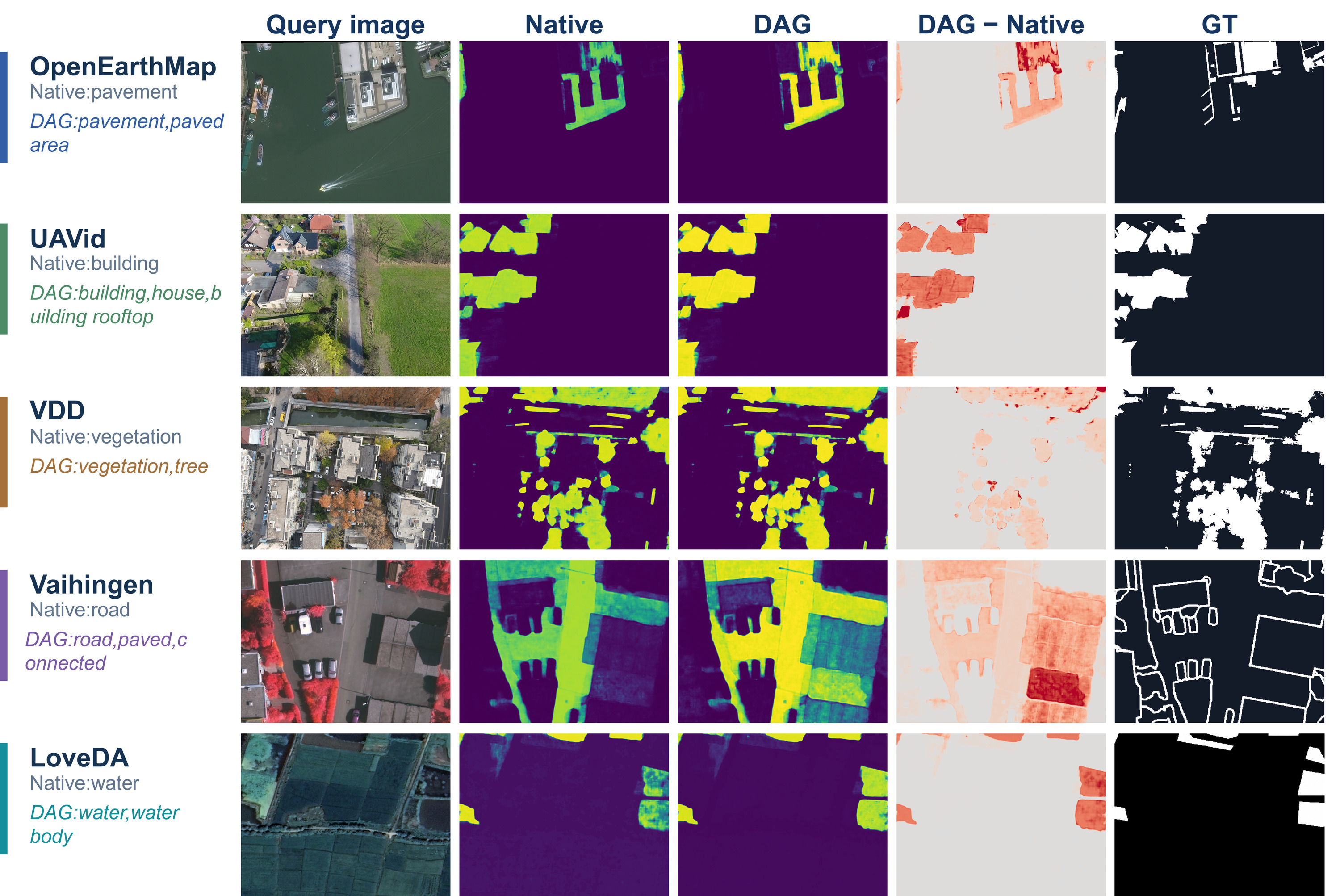}
\par\vspace{-6pt}
\caption{SAM~3 responses from native anchors and the selected
PromptBank. Difference maps highlight regions activated by the
additional descriptions.}
\label{fig:response_visualization}
\end{figure}

Fig.~\ref{fig:response_visualization} compares SAM~3 responses from
native anchors and their DAG-selected PromptBanks. Residual prompts
increase responses over paved regions in OEM and building rooftops
in UAVid, illustrating how additional descriptions can supplement
weak or incomplete native activations. Similar localized changes
appear in vegetation regions in VDD and water regions in LoveDA.
Rather than uniformly raising response scores, the selected prompts
mainly affect particular spatial regions. However, the Vaihingen
example also shows increased activation over a rooftop adjacent
to the intended road region. This illustrates that additional
coverage can be accompanied by responses to competing classes,
motivating the cross-class protection criterion in DAG selection.
Because these maps visualize predictions rather than ground-truth
correctness, they are interpreted as evidence of response behavior,
not as a direct measurement of recovered true-positive pixels.

\paragraph{DAG Selection Criteria}
Table~\ref{tab:dag_criterion_ablation} examines the effects of the
DAG selection criteria on OEM, UDD5, and Vaihingen. Candidate pools,
the frozen SAM~3 backbone, data splits, and evaluation procedures
are held fixed, and SICH is disabled. Full DAG selection achieves
62.8\% mean mIoU, exceeding independent candidate ranking (61.5\%)
by 1.3 points. The improvement is most pronounced on Vaihingen
(65.3\% versus 62.8\%); the corresponding gains on OEM and UDD5
are 0.2 and 1.1 points. This comparison indicates that evaluating
candidate combinations can be more effective than ranking
individual descriptions in isolation, particularly on Vaihingen.

Removing native-anchor protection produces the largest performance
loss in this ablation. Mean mIoU drops from 62.8\% to 43.2\%, a
reduction of 19.6 points. UDD5 decreases from 73.7\% to 29.7\%, and
OEM from 49.3\% to 35.3\%, whereas Vaihingen shows a smaller decline
from 65.3\% to 64.7\%. The strong but uneven effect indicates that
preserving native-anchor evidence is particularly important for the
selected prompt configurations on UDD5 and OEM. These aggregate
results do not, by themselves, establish which classes or regions
account for the loss. In contrast, removing cross-class protection
reduces mean mIoU by 0.5 points to 62.3\%. OEM remains unchanged at
49.3\%, while UDD5 and Vaihingen decrease by 1.0 and 0.3 points,
respectively. Thus, cross-class protection provides a smaller,
dataset-dependent additional benefit when the other selection
criteria remain active.

Disabling fold consistency reduces mean mIoU to 60.1\%, with losses
of 1.4, 2.1, and 4.6 points on OEM, UDD5, and Vaihingen,
respectively. Unlike the cross-class protection ablation, this
variant degrades performance on all three datasets, supporting the
use of agreement across calibration folds to select reusable prompt
sets. The resulting banks also differ in online inference cost.
Full DAG selection achieves 2.52 macro-FPS, compared with 2.38 for
independent ranking. Removing native-anchor protection, cross-class
protection, or fold consistency yields 3.20, 2.59, and 3.44
macro-FPS, respectively. As these measurements are obtained after
offline selection, they characterize the inference cost of the
resulting frozen PromptBanks rather than the computational overhead
of checking individual criteria. Macro-FPS is the arithmetic mean
of the per-dataset inference throughputs.

\begin{table}[!t]
\caption{DAG selection criterion ablation on OEM, UDD5, and Vaihingen
(mIoU, \%). SICH is disabled, and FPS is macro-averaged.}
\label{tab:dag_criterion_ablation}
\centering
\footnotesize
\renewcommand{\arraystretch}{1.05}
\setlength{\tabcolsep}{0.85pt}
\begin{tabularx}{\columnwidth}{@{}>{\raggedright\arraybackslash}Xccccc@{}}
\toprule
Variant & OEM & UDD5 & Vaihingen & Mean & FPS \\
\midrule
Independent ranking
& 49.1 & 72.6 & 62.8 & 61.5 & 2.38 \\
w/o native anchor protection
& 35.3 & 29.7 & 64.7 & 43.2 & 3.2 \\
w/o cross-class protection
& \textbf{49.3} & 72.7 & 65.0 & 62.3 & 2.59 \\
w/o fold consistency
& 47.9 & 71.6 & 60.7 & 60.1 & \textbf{3.44} \\
Full DAG selection
& \textbf{49.3} & \textbf{73.7}
& \textbf{65.3} & \textbf{62.8} & 2.52 \\
\bottomrule
\end{tabularx}
\end{table}

Table~\ref{tab:native_prompt_protection} provides an additional
comparison on the full iSAID split. The full configuration retains
fixed native anchors, enables protection, and selectively includes
residual prompts, reaching 35.1\% mIoU. The unrestricted variant
allows anchor updates, disables protection, and removes restrictions
on the residual set; its performance falls to 20.6\%, a difference
of 14.5 points. Since these changes occur simultaneously, the
comparison evaluates their joint effect and cannot attribute the
entire decrease to native-anchor protection alone. When the anchors
and protection are kept fixed, including every initially accepted
candidate yields 34.2\% mIoU, compared with 35.1\% for selective
inclusion. This more controlled result supports selecting a compact
combination of residual prompts rather than retaining all candidates
that pass individual screening.

\begin{table}[!t]
\caption{Native-anchor protection and residual-prompt inclusion
on the full iSAID split (mIoU, \%).}
\label{tab:native_prompt_protection}
\centering
\footnotesize
\renewcommand{\arraystretch}{1.05}
\setlength{\tabcolsep}{2.5pt}
\begin{tabular*}{\columnwidth}{@{\extracolsep{\fill}}lcccrr@{}}
\toprule
Variant & Anchor & Protection & \makecell{Residual\\set} & mIoU & $\Delta$ \\
\midrule
Full & Fixed & \cmark & Selective & \textbf{35.1} & 0.0 \\
Unrestricted & Updatable & $-$ & Unrestricted & 20.6 & $-14.5$ \\
All accepted & Fixed & \cmark & All accepted & 34.2 & $-0.9$ \\
\bottomrule
\end{tabular*}
\end{table}

\begin{table}[!t]
\caption{Response fusion comparison using the same PromptBank (mIoU, \%).}
\label{tab:fusion_strategy_comparison_four}
\centering
\footnotesize
\renewcommand{\arraystretch}{1.08}
\setlength{\tabcolsep}{3pt}
\begin{tabular*}{\columnwidth}
{@{\extracolsep{\fill}}lrrrrr@{}}
\toprule
Fusion strategy & OEM & Potsdam & Vaihingen & UAVid & Mean \\
\midrule
Semantic only
& 48.0 & 58.5 & 65.1 & 59.6 & 57.8 \\
Instance only
& 28.4 & 47.5 & 57.0 & 56.5 & 47.4 \\
Mean fusion
& 45.9 & 57.4 & 63.4 & 58.6 & 56.3 \\
Max fusion
& \textbf{49.3} & 58.6 & 65.2 & 59.6 & 58.2 \\
SICH (Ours)
& \textbf{49.3} & \textbf{58.9}
& \textbf{65.3} & \textbf{59.9}
& \textbf{58.4} \\
\bottomrule
\end{tabular*}
\end{table}

\begin{table*}[!t]
\caption{Performance under six corruptions, averaged over eight datasets
(mIoU and mAcc, \%).}
\label{tab:corruption_broad}
\centering
\footnotesize
\renewcommand{\arraystretch}{1.05}
\setlength{\tabcolsep}{2.6pt}
\begin{tabular*}{\textwidth}{@{\extracolsep{\fill}}lccccccc@{}}
\toprule
Corruption / metric & VIP & CorrCLIP & RSCLIP & SegEarth-OV & ActiveSAM &
SegEarth-OV3 & \textbf{\method{}} \\
\midrule
Gaussian noise & 31.9 & 22.2 & 34.0 & 33.8 & 48.1 & 48.4 & \textbf{52.4} \\
Motion blur & 32.4 & 24.3 & 36.0 & 36.2 & 50.2 & 51.7 & \textbf{54.4} \\
Downsampling & 32.6 & 24.6 & 36.9 & 37.2 & 50.8 & 52.7 & \textbf{55.5} \\
Fog & 33.8 & 23.7 & 36.7 & 37.0 & 50.9 & 52.5 & \textbf{55.7} \\
Shadow & 33.6 & 24.1 & 34.8 & 34.9 & 49.8 & 53.4 & \textbf{55.8} \\
Stripe artifacts & 32.9 & 23.8 & 36.0 & 36.3 & 50.4 & 52.1 & \textbf{54.2} \\
\midrule
Avg. mIoU & 32.9 & 23.8 & 35.7 & 35.9 & 50.0 & 51.8 & \textbf{54.7} \\
Avg. mAcc & 54.2 & 39.8 & 54.1 & 54.9 & 65.4 & 68.8 & \textbf{72.0} \\
\bottomrule
\end{tabular*}
\end{table*}

\begin{table}[!t]
\caption{Corruption-averaged mIoU (\%) for native anchors and the
frozen PromptBank on four datasets.}
\label{tab:corruption_main}
\centering
\footnotesize
\renewcommand{\arraystretch}{1.15}
\setlength{\tabcolsep}{3pt}
\begin{tabular*}{\columnwidth}{@{\extracolsep{\fill}}l
S[table-format=2.1] S[table-format=2.1] S[table-format=+1.1]@{}}
\toprule
& \multicolumn{2}{c}{Corruption-averaged mIoU (\%)}
& \multicolumn{1}{c}{\multirow{2}{*}{Gain}} \\
\cmidrule(l){2-3}
\multicolumn{1}{l}{Dataset}
& \multicolumn{1}{c}{Native anchors}
& \multicolumn{1}{c}{PromptBank}
& \\
\midrule
OEM & 46.7 & \textbf{48.1} & +1.4 \\
LoveDA       & 40.7 & \textbf{43.4} & +2.7 \\
Vaihingen    & 56.5 & \textbf{61.7} & +5.2 \\
VDD          & 64.1 & \textbf{67.0} & +2.9 \\
\midrule
\textbf{Mean} & \textbf{52.0} & \textbf{55.1} & \textbf{+3.1} \\
\bottomrule
\end{tabular*}
\par\smallskip
\parbox{\columnwidth}{\footnotesize\textit{Note:} Inference speed is 4.2~FPS (Native anchors) and
2.5~FPS (PromptBank) under the same inference setup.}
\end{table}

\begin{figure}[!t]
    \centering
    \subfloat[\label{fig:capacity_miou}]{%
        \includegraphics[
            width=0.49\columnwidth
        ]{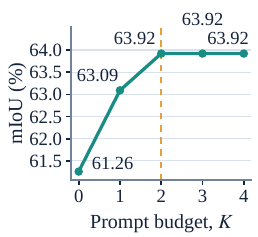}%
    }%
    \hspace{0.015\columnwidth}%
    \subfloat[\label{fig:capacity_fps}]{%
        \includegraphics[
            width=0.49\columnwidth
        ]{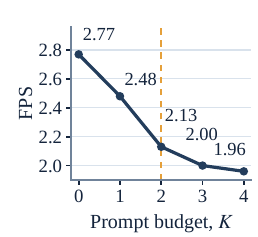}%
    }
    \caption{Effect of the residual-prompt budget on mean mIoU and
inference throughput across OEM, UDD5, Vaihingen, and VDD. Larger
budgets increase inference cost with little gain beyond $K=2$.}
    \label{fig:capacity}
\end{figure}

\paragraph{Prompt Budget}
We examine the maximum number $K$ of residual prompts allowed per
class on OEM, UDD5, Vaihingen, and VDD. As shown in
Fig.~\ref{fig:capacity}(a), mean mIoU increases with the initial
budget increments and reaches approximately 63.9\% at $K=2$.
Increasing $K$ further produces little additional improvement
through $K=4$, indicating diminishing returns from larger prompt
sets within the evaluated candidate pools. Meanwhile, throughput
decreases from 2.13 FPS at $K=2$ to 1.96 FPS at $K=4$, as shown in
Fig.~\ref{fig:capacity}(b). The additional responses therefore
increase online computation without a comparable mean accuracy
gain beyond $K=2$. We adopt this budget as a practical compromise
for the tested datasets and candidate pools; the experiment does
not establish a universally optimal budget.

\paragraph{Response Fusion Strategies}
Table~\ref{tab:fusion_strategy_comparison_four} compares response
fusion strategies using the same frozen PromptBank and matched
SAM~3 responses on OEM, Potsdam, Vaihingen, and UAVid. Semantic-only
inference attains 57.8\% mean mIoU, compared with 47.4\% for
instance-only inference. The gap is especially large on OEM
(48.0\% versus 28.4\%) and Potsdam (58.5\% versus 47.5\%), while
instance-only inference is closer to semantic-only inference on
UAVid (56.5\% versus 59.6\%). These differences indicate that
instance responses do not provide equally reliable class evidence
across the tested datasets. Simple mean fusion achieves 56.3\%,
below semantic-only inference, whereas max fusion raises the mean
to 58.2\%. SICH achieves 58.4\%, matching max fusion on OEM and
improving Potsdam, Vaihingen, and UAVid by 0.3, 0.1, and 0.3
points, respectively. Although these gains are modest, the results
suggest that selective response corrections can offer a small
additional benefit over direct max fusion without changing the
selected prompts. Together with Table~\ref{tab:component_ablation},
this comparison identifies DAG-selected prompts as the main source
of improvement and SICH as a complementary refinement.

\begin{figure}[!t]
    \centering
    \includegraphics[width=0.48\textwidth]{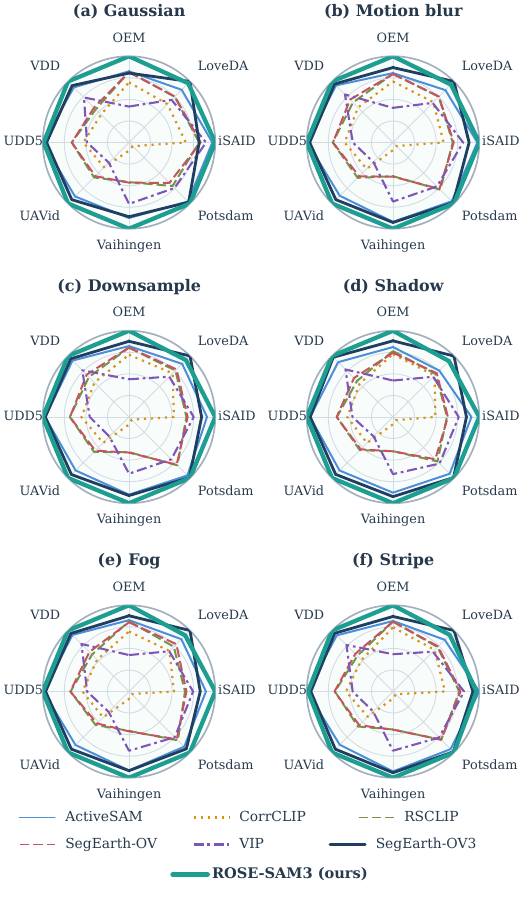}
    \par\vspace{-5pt}
    \caption{Dataset-level mIoU across six RS-OVSS-C corruptions.
    Each panel compares methods on the same eight datasets. Axis
    bounds are fixed across panels for each dataset.}
    \label{fig:corruption_radar}
\end{figure}

\begin{figure*}[!t]
\centering
\includegraphics[width=\linewidth,keepaspectratio]{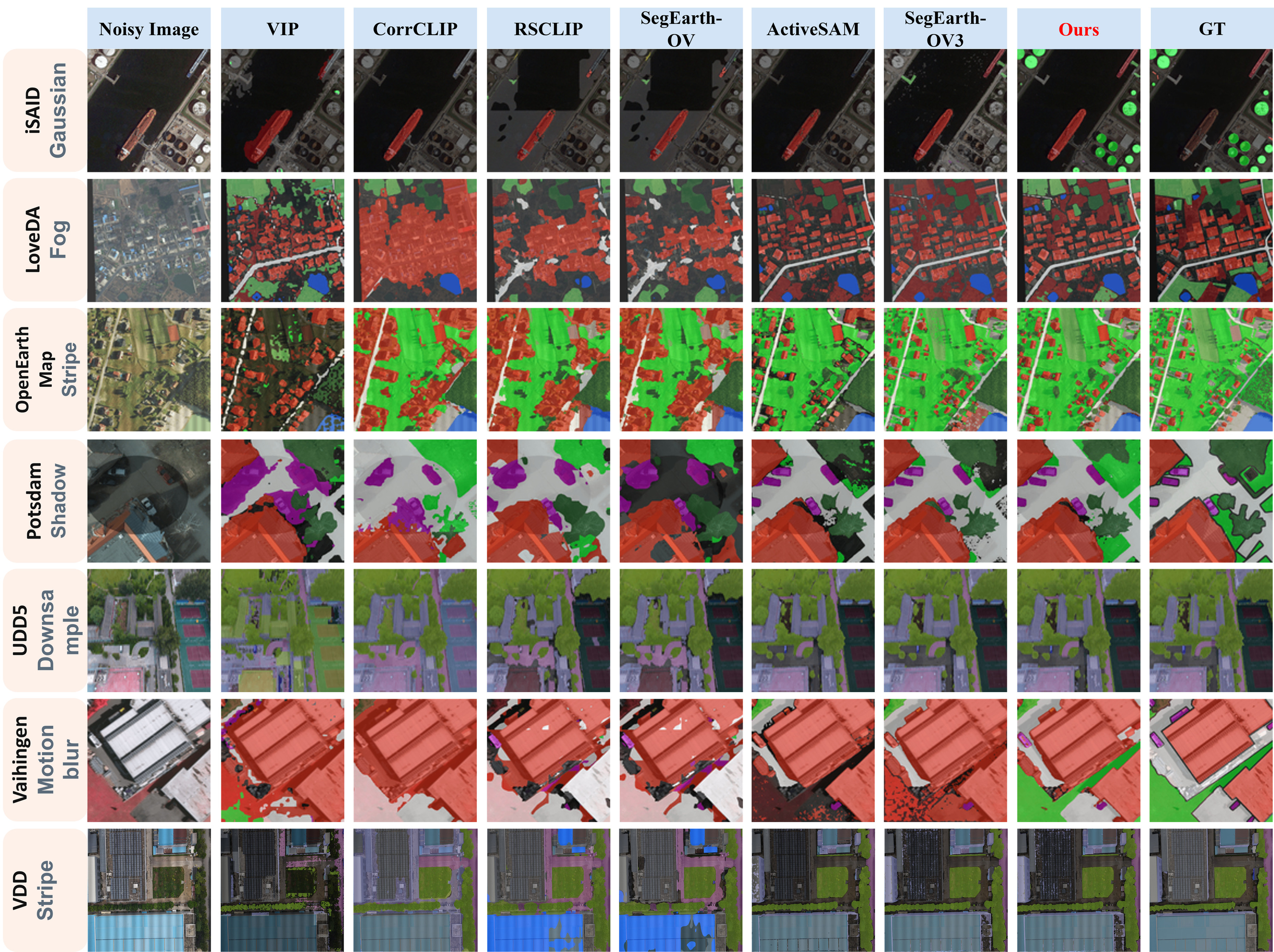}
\par\vspace{-6pt}
\caption{Qualitative comparison under RS-OVSS-C corruptions.
The examples contrast preservation of broad regions with errors
around small objects and fine boundaries.}
\label{fig:qualitative_noisy}
\end{figure*}

\subsection{Robustness under Image Corruptions}

We evaluate performance under image degradation using RS-OVSS-C. Table~\ref{tab:corruption_broad} compares seven methods across six corruption types and eight datasets. ROSE-SAM3 uses the PromptBanks selected on clean images without reselection. It achieves the highest dataset-averaged mIoU for each corruption type, with an overall mIoU of 54.7\% and mAcc of 72.0\%, compared with 51.8\% and 68.8\% for SegEarth-OV3. Its lead is largest under Gaussian noise (+4.0 points) and smallest under stripe artifacts (+2.1). Gaussian noise changes local pixel values, while the stripe setting introduces structured intensity variation. The larger margin under noise is consistent with selected descriptions supplying additional class cues when local appearance is less reliable, although the benchmark does not isolate this mechanism. Dataset-level results in Fig.~\ref{fig:corruption_radar} show that ROSE-SAM3 ranks highest in 39 of the 48 dataset--corruption combinations. Its advantage is particularly evident on iSAID, Vaihingen, and UAVid, while SegEarth-OV3 remains competitive on LoveDA and ActiveSAM leads in some cases, showing that the average lead does not imply uniform robustness across datasets.

We next test whether residual prompts selected on clean images remain useful under corruption. Table~\ref{tab:corruption_main} compares native anchors with the frozen PromptBank under the same inference procedure on OEM, LoveDA, Vaihingen, and VDD. Averaged over these four datasets and six corruptions, mIoU increases from 52.0\% to 55.1\%, with gains on every dataset. The gain is largest on Vaihingen (+5.2), followed by VDD (+2.9), LoveDA (+2.7), and OEM (+1.4). This spread suggests that the benefit depends on how well the added descriptions cover each dataset's classes and appearances. The qualitative examples provide concrete cases: under fog, building regions in LoveDA remain more complete; with stripe artifacts, the OEM scene layout and road connections remain more continuous. In contrast, shadow and motion blur still weaken boundaries and thin structures, so added prompts do not restore spatial detail removed or obscured in the image. Gains occur under all six corruptions, ranging from 2.5 points under fog to 3.6 under Gaussian noise. Inference throughput decreases from 4.2 to 2.5 FPS because the PromptBank requires responses for additional prompts. This comparison measures accuracy under corruption, rather than the drop from clean-image accuracy. Geographic and sensor shifts are outside this evaluation.

Fig.~\ref{fig:qualitative_noisy} presents qualitative results
under image corruptions. Under Gaussian noise in iSAID, ROSE-SAM3
recovers more small circular target regions
than the competing methods. Under fog in LoveDA, it better separates
neighboring buildings and reduces fragmented predictions.
For OEM with stripe artifacts, it preserves the main
vegetation and built-up regions while reducing class confusion.
Under shadow in Potsdam, the main building structures remain
identifiable, although some boundary errors persist.
For UDD5 under downsampling, the overall spatial arrangement
of roads and vegetation is largely preserved.
Under motion blur in Vaihingen, ROSE-SAM3 produces a more
complete building region with fewer surrounding
misclassifications. In VDD with stripe artifacts, the main
rooftop and vegetation regions remain consistent with the
ground truth, although local boundary errors remain.
These observations are consistent with
Tables~\ref{tab:corruption_broad} and~\ref{tab:corruption_main}:
ROSE-SAM3 maintains competitive performance under image
corruptions, while the selected PromptBank improves over
native anchors on the evaluated datasets.

\section{Conclusion}

We presented ROSE-SAM3, a training-free framework for open-vocabulary remote-sensing segmentation. It selects complementary descriptions using frozen SAM~3 responses on unlabeled images. Offline DAG search evaluates prompt combinations, and SICH refines their semantic responses at inference. We also introduced RS-OVSS-C, which applies six controlled corruption families to eight remote-sensing datasets while preserving class vocabularies and reference labels. ROSE-SAM3 achieves 57.3\% mean mIoU on the clean benchmarks and 54.7\% on RS-OVSS-C, the highest averages among the compared training-free methods under the reported protocols. Ablations assess DAG selection and SICH, and the selected prompts remain useful on corrupted images. These results support model responses as a basis for selecting prompts without updating model parameters.




\end{document}